\documentclass[letterpaper, 10 pt, conference]{ieeeconf}  

\IEEEoverridecommandlockouts                              

\usepackage{cite}
\usepackage{svg}
\usepackage{amsmath,amssymb,amsfonts}
\usepackage{algorithmic}
\usepackage{graphicx}
\usepackage{textcomp}
\usepackage{xcolor}
\usepackage{dsfont}
\usepackage{subfigure}
\usepackage{booktabs}
\usepackage{bbm}
\usepackage{url}
\usepackage{svg}
\usepackage{multirow}
\usepackage{multicol}
\usepackage{rotating}
\usepackage{color}
\usepackage{bigstrut}
\usepackage{array}
\usepackage{tabularx}
\usepackage{amssymb}
\usepackage{pifont}

\usepackage[normalem]{ulem}
\useunder{\uline}{\ul}{}
\usepackage{marvosym}

\makeatletter
\let\NAT@parse\undefined
\makeatother
\usepackage{hyperref}
\usepackage[nameinlink,noabbrev]{cleveref}

\title{\LARGE \bf
LODE: Locally Conditioned Eikonal Implicit Scene Completion from Sparse LiDAR
}

\author{Pengfei Li$^{1,2}$, Ruowen Zhao$^{1,3}$, Yongliang Shi$^{1}$, Hao Zhao$^{1}$, Jirui Yuan$^{1}$, Guyue Zhou$^{1}$\textsuperscript{\Letter} and Ya-Qin Zhang$^{1}$
\thanks{$^{1}$ Institute for AI Industry Research (AIR), Tsinghua University, China
        \{Shiyongliang,zhaohao,yuanjirui,zhouguyue,zhangyaqin\}@air.tsinghua.edu.cn.}%
\thanks{$^{2}$Department of Computer Science and Technology, Tsinghua University, China,
        li-pf22@mails.tsinghua.edu.cn.}%
\thanks{$^{3}$ University of Chinese Academy of Sciences, China
        zhaorewen20@mails.ucas.ac.cn.}%
}

\begin{document}

\maketitle
\thispagestyle{empty}
\pagestyle{empty}

\begin{abstract}

Scene completion refers to obtaining dense scene representation from an incomplete perception of complex 3D scenes. 
This helps robots detect multi-scale obstacles and analyse object occlusions in scenarios such as autonomous driving.
Recent advances show that implicit representation learning can be leveraged for continuous scene completion and achieved through physical constraints like Eikonal equations.
However, former Eikonal completion methods only demonstrate results on watertight meshes at a scale of tens of meshes.
None of them are successfully done for non-watertight LiDAR point clouds of open large scenes at a scale of thousands of scenes.
In this paper, we propose a novel Eikonal formulation that conditions the implicit representation on localized shape priors which function as dense boundary value constraints, and demonstrate it works on SemanticKITTI and SemanticPOSS.
It can also be extended to semantic Eikonal scene completion with only small modifications to the network architecture.
With extensive quantitative and qualitative results, we demonstrate the benefits and drawbacks of existing Eikonal methods, which naturally leads to the new locally conditioned formulation. Notably, we improve IoU from 31.7\% to 51.2\% on SemanticKITTI and from 40.5\% to 48.7\% on SemanticPOSS.
We extensively ablate our methods and demonstrate that the proposed formulation is robust to a wide spectrum of implementation hyper-parameters.
Codes and models are publicly available at \url{https://github.com/AIR-DISCOVER/LODE}.

\end{abstract}


\section{Introduction}

Representing 3D data with neural implicit functions is actively explored recently due to its strong modeling capability and memory efficiency \cite{c17, c18, c19, c14, lee2019online, c10, duan2020curriculum}. 
Meanwhile, it can be easily meshed and rendered to facilitate human viewing.
While most methods are fully supervised \cite{c17, c18, c19}, SIREN \cite{c10} proposes an Eikonal implicit scene completion method with weak supervision needed.
It learns a signed distance function (SDF), which measures the nearest distance to the scene surface, through the process of solving an Eikonal differential equation with only points on the surface and without knowing SDF values in free space.
This scheme is promising for large-scale sparse LiDAR data because (1) for non-water-tight scenes, it is hard to define signed distance in free space and (2) it requires less supervision than non-Eikonal formulations and thus is simpler, especially when considering the completion of thousands of scenes.

\begin{figure}[tpb]
\centerline{\includegraphics[width=0.5\textwidth]{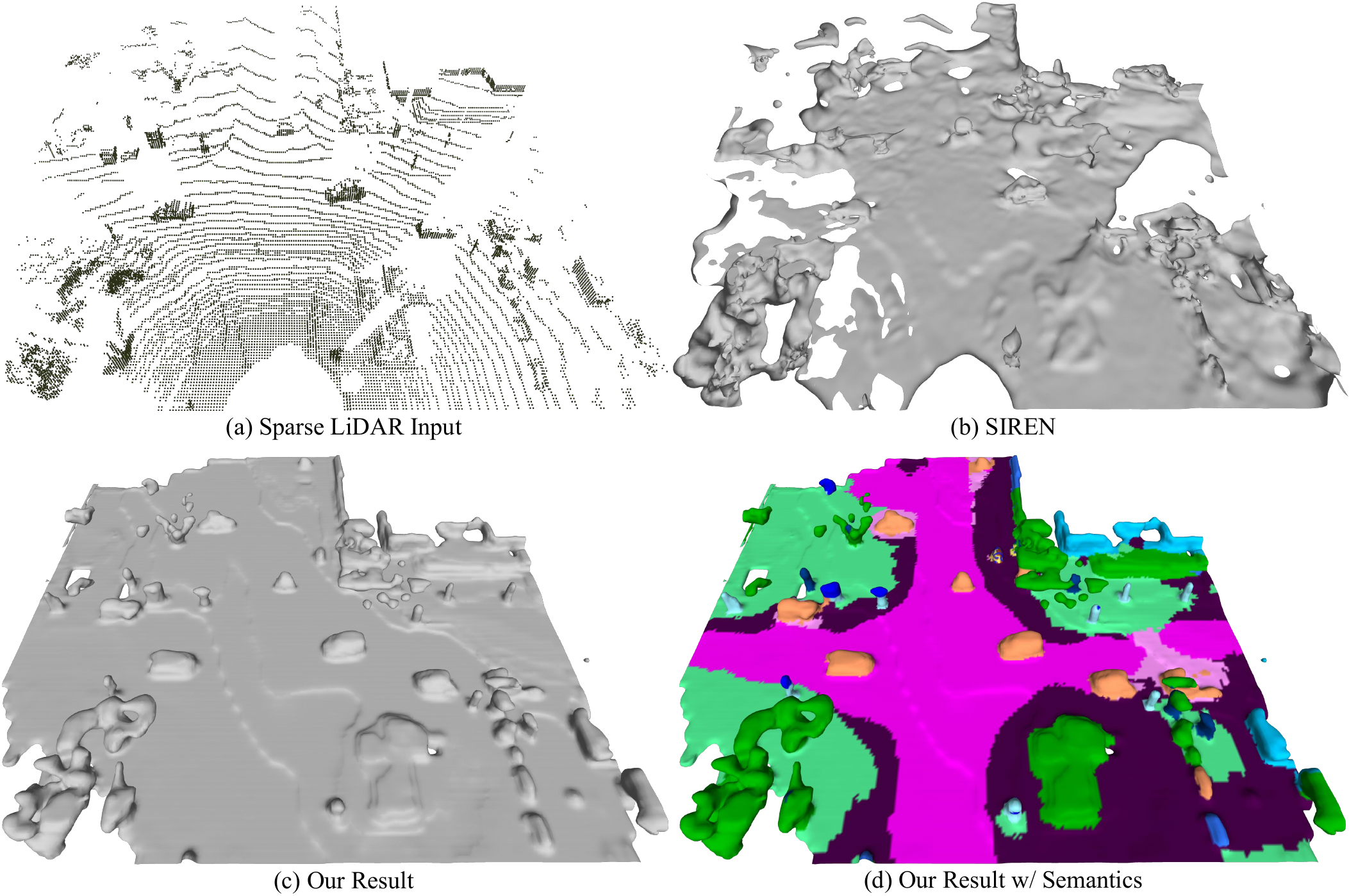}}
\caption{(a) The input is a sparsity-variant point cloud of road scenes captured by LiDAR. (b) The implicit fitting result of SIREN \cite{c10}. Note that this is well tuned by an exhaustive parameter search. (c) The output of our method is a neural signed distance function of arbitrary resolution, i.e., implicit scene completion. (d) Our result with extended semantic parsing.}
\label{fig:teaser}
\end{figure}

However, even after an exhaustive parameter search, SIREN fails to fit sparse LiDAR data (Fig.~\ref{fig:teaser}-a), which limits its application in many important scenarios such as autonomous driving. This is understandable as SIREN is a pure generative model and LiDAR point clouds are extremely sparse. Specifically, the reasons are three-fold: (1) The sparsity of on-surface points amplifies the negative impact of wrongly sampled off-surface anchors. (2) The normal orientations of sparse points cannot be estimated accurately from their neighbors, which serve as a necessary boundary value constraint for SIREN fitting. (3) Without trustworthy boundary values, enforcing a hard Eikonal constraint leads to even inaccurate SDF values. As shown in Fig.~\ref{fig:teaser}-b, the SIREN fitting result is fragmented.

To overcome these limitations, we develop a novel Eikonal implicit formulation by introducing an intermediate embedding domain, where localized shape priors are contained. 
Instead of directly fitting a function to map 3D Cartesian coordinates to signed distances, we first map the Euclidean space to a corresponding high-dimensional shape embedding space, and then the signed distance space.
These shape embeddings function as dense boundary values that entangle both zeroth-order (on-surface points) and first-order (normal directions) constraints, in a data-driven manner.
Naturally, the issue of enforcing a hard Eikonal constraint is also alleviated. 
This proposed formulation is named \textbf{L}ocally C\textbf{o}n\textbf{d}itioned \textbf{E}ikonal Formulation and abbreviated as \textbf{LODE}.
The result of LODE is significantly better than SIREN (Fig.~\ref{fig:teaser}-c).
And the supplementary video demonstrates LODE performs well on in-the-wild sequential LIDAR inputs.

Specifically, to implement LODE, we propose a novel hybrid architecture combining a discriminative model and a generative model.
The discriminative part of our method exploits the strong representation learning power of sparse convolution, generating latent shape embeddings from sparse point cloud input.
The generative model takes as input the ground truth point cloud coordinates along with pointwise latent shape embeddings retrieved by trilinear sampling and predicts SDF values of these points. 
During inference, the ground truth points are replaced with the points of interest.

Furthermore, to demonstrate the flexibility of LODE, we extend our method to implicit semantic completion in two ways: (1) by adding a dense discriminative head to predict semantic labels which can be mapped to the implicit function using K-Nearest-Neighbors; (2) by adding a parallel implicit generative head to directly model the implicit semantic field. We evaluate them on SemanticKITTI and achieve results (Fig.~\ref{fig:teaser}-d) comparable to state-of-the-art methods.

To summarize, our contributions are as follows:

\begin{itemize}
\item[$\bullet$] We develop a locally conditioned Eikonal implicit scene completion formulation that incorporates learned shape priors as dense boundary value constraints. 
\item[$\bullet$] We apply the formulation in road scene understanding, leading to the first Eikonal implicit road scene completion method without knowing SDF values in free space.
\item[$\bullet$] We achieve state-of-the-art completion results on SemanticKITTI and SemanticPOSS, outperforming the best Eikonal completion results by +19.5\% and +8.2\% IoU. Code, data, and models will be released.
\end{itemize}

\section{Related Works}
\textbf{Neural Implicit Representation.} 
The general principle of neural implicit representation is to train a neural network to approximate a continuous function that is hard to parameterize otherwise. \cite{c17} proposes to learn deep signed distance functions conditioned on shape embeddings. \cite{c18} approximates occupancy functions with conditional batchnorm networks. \cite{c13} introduces data-driven shape embeddings into occupancy networks for indoor scene completion. \cite{c19} uses hyperplanes as compact implicit representations to reconstruct shapes sharply and compactly.
\cite{c10} shows that using gradient supervision allows Eikonal SDF learning with only on-surface SDF value supervision and sine activations are critical to its success.
\cite{c29} combines Gaussian ellipsoids and implicit residuals to represent shapes accurately. Some recent works exploit 3D implicit representations for instance-level understanding from point cloud \cite{c30} or RGB \cite{c22} inputs.
Despite these advances, there is no work yet on implicit scene completion on LiDAR point clouds where the data is extremely sparse with heterogeneous distribution.
Our method bridges this gap with the proposed locally conditioned Eikonal formulation.

\textbf{LiDAR-based scene understanding.} While there are many advances in camera-based cognitive scene understanding \cite{chuang2018learning, chen2022cerberus, ye2017can, li2022toist, li2022distance}, LiDAR point cloud provides reliably accurate 3D structural information and thus has been mainly leveraged in geometric scene understanding.
Numerous LiDAR-based SLAM methods are proposed to improve the quality and real-time performance \cite{hess2016real, droeschel2018efficient, vizzo2021poisson, pan2021mulls, ramezani2020online}. \cite{bowman2017probabilistic, chen2019suma++, li2021sa} further incorporate semantic understanding into LiDAR SLAM systems. 
These methods explicitly complete a scene with multiple frames of LiDAR data, while our goal is to achieve implicit completion with a single frame.
Sparse LiDAR points are also efficiently leveraged in depth completion \cite{ma2019self, choi2021stereo}, object detection \cite{maturana20153d, dou2019seg}, segmentation \cite{chen2021moving, li2021rethinking, yi2021complete, milioto2020lidar, li2021multi, behley2021benchmark}.
We believe the performance of these methods will be boosted with LODE completing the original sparse representation.
Moreover, by providing fine geometric details, LODE may aid in the detection of anomalous obstacles \cite{asvadi20163d} and some other tasks.

\section{Formulation}

\begin{figure*}[t]
\centerline{\includegraphics[width=0.9\textwidth]{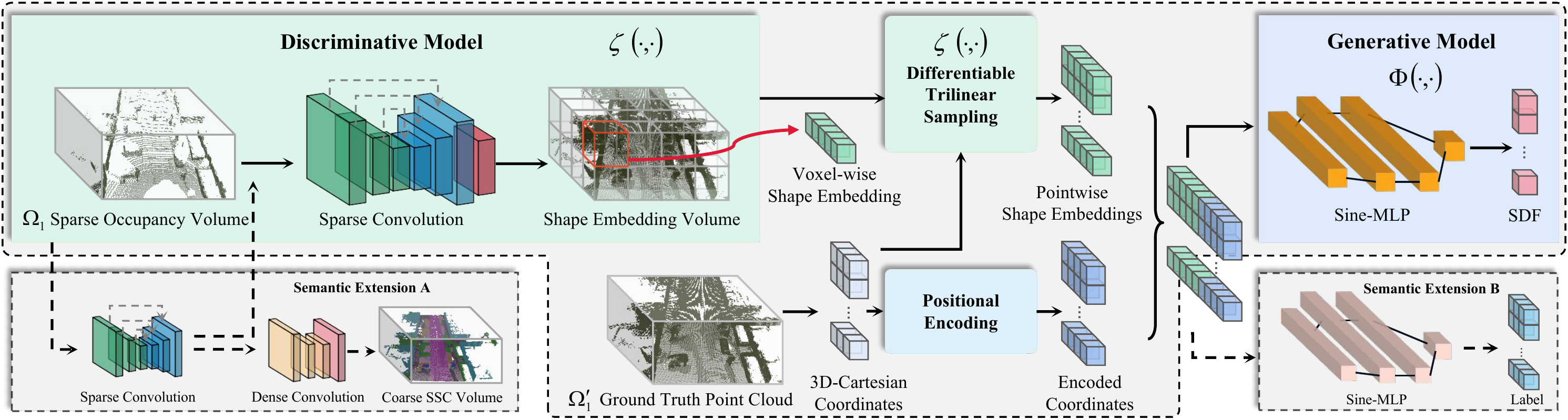}}
\caption{\textbf{Overview of our architecture.} The discriminative model extracts shape priors and the generative model predicts SDF values. They are bridged by differentiable triliner sampling. Positional Encoding is used to represent more details. Two semantic extension options are outlined in small dashed boxes.}
\label{fig:mainfig}
\vspace{-0.5cm}
\end{figure*}

\subsection{Eikonal Implicit Completion Formulation}

Eikonal completion methods aim at fitting the signed distance function (SDF) of a scene. The signed distance is the nearest distance from a point of interest to the scene surface, with the sign denoting whether the point is located outside (positive) or inside (negative) of the surface. The iso-surface where the signed distance equals zero implicitly delineates the scene. Formally, the goal of Eikonal implicit completion is to find a function $\Phi(\textbf{x})$, which satisfies a set of $M$ constraints $\mathcal{C}_m$, to approximate the underlying SDF. Each constraint relates the function $\Phi(\textbf{x})$ or its gradient to certain input quantities
$\textbf{a}(\textbf{x})$ on the corresponding domain $\Omega_m$:
\begin{equation}
\begin{split}
\mathcal{C}_m (\textbf{a}(\textbf{x}),&\Phi(\textbf{x}),\nabla_\textbf{x}\Phi(\textbf{x})) = 0, \\
&\forall \textbf{x} \in \Omega_m, m=0,...,M-1.
\end{split}
\end{equation}
Specifically, these constraints are required:
\begin{equation}
\mathcal{C}_0 := |\nabla_\textbf{x}\Phi(\textbf{x})| - 1, \textbf{x} \in \Omega_0.
\end{equation}
\begin{equation}
\mathcal{C}_1 := \nabla_\textbf{x}\Phi(\textbf{x}) - \textbf n(\textbf x), \textbf{x} \in \Omega_1.
\end{equation}
\begin{equation}
\mathcal{C}_2 := \Phi(\textbf{x}) - \rm{SDF}(\textbf{x}), \textbf{x} \in \Omega_2.
\end{equation}
Here, $\mathcal{C}_0$ guarantees $\Phi(\textbf{x})$ satisfies the Eikonal equation in the whole physical space of interest ($\Omega_0$), which is a intrinsic property of SDF. $\mathcal{C}_1$ forces that the gradients of $\Phi(\textbf{x})$ equal the normal vectors for input on-surface points ($\Omega_1$). $\mathcal{C}_2$ constrains the values of $\Phi(\textbf{x})$ equal the ground truth SDF for labeled anchor points ($\Omega_2$). In this way, the problem can be regarded as an Eikonal boundary value problem, where the differential equation $\mathcal{C}_0$ is to be solved under the first-order constraint $\mathcal{C}_1$ and the zeroth-order constraint $\mathcal{C}_2$.

However, the ground truth SDF values in free space are difficult to obtain. A recent method named SIREN \cite{c10} proposes an intriguing variant where the domain of $\mathcal{C}_2$ is limited to on-surface points in $\Omega_1$. As the ground truth SDF values of points in $\Omega_1$ are zero, $\mathcal{C}_2$ is reduced to:
\begin{equation}
\mathcal{C}_2 := \Phi(\textbf{x}), \textbf{x} \in \Omega_1.
\end{equation}
To remedy the lack of constraints on off-surface points, SIREN introduces another constraint:
\begin{equation}
\mathcal{C}_3 := \psi(\Phi(\textbf{x})), \textbf{x} \in \Omega_3.
\end{equation}
Here, $\psi$ pushes $\Phi(\textbf{x})$ values away from 0, for randomly and uniformly sampled off-surface points ($\Omega_3 \subseteq \Omega_0\setminus\Omega_1$).

Nevertheless, these constraints fail to cope with the scenario where on-surface points in $\Omega_1$ are sampled from sparse LiDAR point cloud data. Reasons are three-fold: (1) The sparsity of on-surface points in $\Omega_1$ amplifies the negative impact of $\mathcal{C}_3$ on the wrongly sampled off-surface anchors in $\Omega_3$ (i.e., located on or near the surface). (2) The normal orientations of sparse points in $\Omega_1$ cannot be estimated accurately from their neighbors, leading to an incorrect constraint $\mathcal{C}_1$. (3) Without trustworthy boundary value constraints $\mathcal{C}_3$ and $\mathcal{C}_1$, enforcing the hard Eikonal constraint $\mathcal{C}_0$ leads to even inaccurate SDF values in free space.

\subsection{Locally Conditioned Eikonal Formulation (LODE)}

To overcome the aforementioned limitations, we propose a locally conditioned Eikonal formulation $\Phi(\textbf{x},\textbf{e})|_{\textbf{e}=\zeta(\textbf{x}, \Omega_1)}$ to approximate SDF. Here, we use $\zeta(\cdot, \cdot)$ to first map the Euclidean space to a high-dimensional shape embedding space. It functions as a dense boundary value constraint for the differential equation. Then $\Phi(\cdot, \cdot)$ maps the shape embedding space to the signed distance space. As a result, the constraints to be satisfied are formally re-written as:
\begin{equation}
\mathcal{C}_0^\prime := |\nabla_\textbf{x}\Phi(\textbf{x},\textbf{e})|_{\textbf{e}=\zeta(\textbf{x}, \Omega_1)}| - 1, \textbf{x} \in \Omega_0.
\end{equation}
\begin{equation}
\mathcal{C}_4 := \rho(\zeta(\textbf{x}, \Omega_1)), \textbf{x} \in \Omega_0.
\end{equation}
We use $\rho(\zeta(\textbf{x}, \Omega_1))$ to represent the underlying dense constraint contained in the shape embedding space, which implicitly entangles correct $\mathcal{C}_1$, $\mathcal{C}_2$, and $\mathcal{C}_3$ constraints of the modified formulations:
\begin{equation}
\mathcal{C}_1^\prime := \nabla_\textbf{x}\Phi(\textbf{x},\textbf{e})|_{\textbf{e}=\zeta(\textbf{x}, \Omega_1)} - \textbf n(\textbf x), \textbf{x} \in \Omega_1^\prime.
\end{equation}
\begin{equation}
\mathcal{C}_2^\prime := \Phi(\textbf{x},\textbf{e})|_{\textbf{e}=\zeta(\textbf{x}, \Omega_1)}, \textbf{x} \in \Omega_1^\prime.
\end{equation}
\begin{equation}
\mathcal{C}_3^\prime := \psi(\Phi(\textbf{x},\textbf{e})|_{\textbf{e}=\zeta(\textbf{x}, \Omega_1)}), \textbf{x} \in \Omega_3^\prime.
\end{equation}
Here, $\Omega_1^\prime$ contains the dense ground truth on-surface points and $\Omega_3^\prime \subseteq \Omega_0\setminus\Omega_1^\prime$. Hence the aforementioned problem of trustworthy boundary values is resolved. Naturally, the issue of enforcing a hard Eikonal constraint is also alleviated.

We implement the proposed LODE formulation in a data-driven manner. The acquisition of functions $\zeta(\cdot, \cdot)$ and $\Phi(\cdot,\cdot)$ can be cast in a loss function that penalizes deviations from the constraints $\mathcal{C}_0^\prime$, $\mathcal{C}_1^\prime$, $\mathcal{C}_2^\prime$, and $\mathcal{C}_3^\prime$ on
their domain:
\begin{equation}
\begin{split}
\mathcal{L}_{\rm{LODE}} 
&= \lambda_1\int_{\Omega_0}  \left\| |\nabla_{\textbf{x}} \Phi(\textbf{x},\textbf{e})|_{\textbf{e}=\zeta(\textbf{x}, \Omega_1)}| -1 \right\|d{\textbf{x}} \\
&+ \lambda_2\int_{\Omega_1^\prime} (1-\langle \nabla_{\textbf{x}} \Phi(\textbf{x},\textbf{e})|_{\textbf{e}=\zeta(\textbf{x}, \Omega_1)}, \textbf{n}(\textbf{x}) \rangle)d{\textbf{x}} \\
&+ \lambda_3\int_{\Omega_1^\prime} \left\| \Phi(\textbf{x},\textbf{e})|_{\textbf{e}=\zeta(\textbf{x}, \Omega_1)}\right\|d{\textbf{x}} \\
&+ \lambda_4\int_{\Omega_3^\prime} \psi(\Phi(\textbf{x},\textbf{e})|_{\textbf{e}=\zeta(\textbf{x}, \Omega_1)})d{\textbf{x}},
\label{equ:sdfloss}
\end{split}
\end{equation}
where $\lambda_1$ - $\lambda_4$ are constant weight parameters and $\langle \cdot,\cdot \rangle$ calculates cosine similarity between two vectors. 


\section{Method}

To realize LODE, we propose a hybrid neural network architecture combining a discriminative model with a generative model, as shown in Fig.~\ref{fig:mainfig}.
The discriminative part
exploits the strong representation learning power of sparse convolution, generating latent shape embeddings from sparse input $\Omega_1$.
It together with the differentiable trilinear sampling module works as function $\zeta(\cdot, \cdot)$. 
The generative model consists of an MLP, functioning as $\Phi(\cdot,\cdot)$.
It takes as input the encoded coordinates of ground truth points $\Omega_1^\prime$ along with pointwise latent shape embeddings and predicts SDF values of these points. Using gradient descent, we can get the optimized $\zeta(\cdot, \cdot)$ and $\Phi(\cdot,\cdot)$ in the parameterized form.

\subsection{Discriminative Model}
\label{subsection:D_Model}
Intuitively, road scenes have the characteristic of repetition. Thus convolutional neural network can be employed as the discriminative model to exploit the translation invariance.

Taking LiDAR points $\Omega_1$ as input, we first conduct voxelization to obtain 3D occupancy volume $V_{\rm{occ}}$ with size $1 \times D_{\rm{occ}} \times W_{\rm{occ}} \times H_{\rm{occ}}$.
Then the discriminative model maps it into a shape embedding volume $V_{\rm{se}}$ with size $d_{\rm{se}} \times D_{\rm{se}} \times W_{\rm{se}} \times H_{\rm{se}}$, where $d_{\rm{se}}$ is the dimension of the shape embedding outputs.
To tackle the sparsity of $V_{\rm{occ}}$, we employ the sparse operations of the Minkowski Engine \cite{c12} to build the model, which is a multiscale encoder-decoder network.
It extracts shape priors via a shape completion process: the encoder consisting of convolutional blocks aggregates localized features, and the decoder involving generative deconvolutional blocks generates dense results.
Yet the constant generation of new voxels will destroy the sparsity just as the \emph{submanifold dilation problem} \cite{c11}.
To avoid this, we use a pruning block to prune off redundant voxels.
It contains a convolutional layer to determine the binary classification result of whether a voxel should be pruned, which is supervised with binary cross-entropy loss:
\begin{equation}
\begin{split}
\mathcal{L}_{\rm{com}} = -{\frac 1 m}\sum_{i=1}^m & {\frac 1 {n_i}} \sum_{j=1} ^ {n_i} [y_{i,j} {\rm{log}}(p_{i,j}) \\
&+ (1-y_{i,j}){\rm{log}}(1-p_{i,j})],
\end{split}
\end{equation}
where $m$ is the count of supervised blocks, $n_i$ denotes the count of voxels in the $i$-th block, $y_{i,j}$ and $p_{i,j}$ are the true and predicted existence probabilities for voxel $i$ respectively.


\subsection{Differentiable Trilinear Sampling Module}

After generating $V_{\rm{se}}$, pointwise shape embedding $\textbf{e}_i \in \mathbb{R}^{d_{\rm{se}}}$ for query point $\textbf{x}_i \in \Omega_0$ is needed.
We use trilinear interpolation to sample $\textbf{e}_i$ for $\textbf{x}_i$ from its 8 nearest voxel centers to maintain the continuity of the latent shape field at the voxel borders.
Formally, with the length of voxel edge normalized, the trilinear sampling for $\textbf{e}_i$ can be written as:
\begin{equation}
\begin{split}
&e_i^c = \sum\limits_{m} ^{D_{\rm{se}}} \sum\limits_{n} ^{W_{\rm{se}}} \sum\limits_{k} ^{H_{\rm{se}}} e_{mnk}^c \times {\rm{max}}(0,1-|x_i-x_m|)\\&\times{\rm{max}}(0,1-|y_i-y_n|)\times{\rm{max}}(0,1-|z_i-z_k|),
\end{split}
\end{equation}
where $e_i^c$ and $e_{mnk}^c$ are shape embeddings on channel $c$ for $\textbf{x}_i=(x_i, y_i, z_i) $ and voxel center $\textbf{x}_{mnk}=(x_m, y_n, z_k)$.
Then the gradient with respect to $\textbf{e}_{mnk}$ for backpropagation is:
\begin{equation}
\begin{split}
&\frac{\partial e_i^c}{\partial  e_{mnk}^c} = \sum\limits_{m} ^{D_{\rm{se}}} \sum\limits_{n} ^{W_{\rm{se}}} \sum\limits_{k} ^{H_{\rm{se}}} {\rm{max}}(0,1-|x_i-x_m|)\\&\times{\rm{max}}(0,1-|y_i-y_n|)\times{\rm{max}}(0,1-|z_i-z_k|).
\end{split}
\end{equation}

This differentiable trilinear sampling mechanism allows loss gradients to flow back to $V_{\rm{se}}$ and further back to the discriminative model, making it possible to train discriminative model and the following generative model cooperatively.


\subsection{Positional Encoding Module}
Positional encoding has proved an effective technique in neural rendering\cite{c14} \cite{c34} for its capacity to capture high-frequency information.
Thus we leverage it to represent more geometric details of the signed distance field.
Specifically, the 3D Cartesian coordinate $\textbf{x}_i$ is encoded into high-dimensional feature $\textbf{y}_i = (\gamma_{\rm{enc}}(x_i), \gamma_{\rm{enc}}(y_i), \gamma_{\rm{enc}}(z_i)) \in \mathbb{R}^{d_{\rm{enc}}}$, where
\begin{equation}
\begin{split}
\gamma_{\rm{enc}}(p) = ({\rm{sin}}(2^0 \pi p), &{\rm{cos}}(2^0 \pi p), \cdots, \\
 &{\rm{sin}}(2^{L-1} \pi p), {\rm{cos}}(2^{L-1} \pi p)).
 \end{split}
\end{equation}
$L$ is the number of frequency octaves and thus $d_{\rm{enc}}=6L$.


\subsection{Generative Model}
We use an MLP as the generative model for implicit SDF representation and use sine as a periodic activation function to better model details \cite{c10}. Thus $\Phi$ can be formalized as:
\begin{equation}
\begin{split}
\Phi(\textbf{x}) = &{\mathbf W}_n(\phi_{n-1} \circ \phi_{n-2} \circ \cdots \circ \phi_0)(\textbf{x})+{\mathbf b}_n, \\
&{\textbf{x}}_j \mapsto \phi_{j}({\textbf{x}}_j)={\rm{sin}}({\mathbf W}_j {\textbf{x}}_j + {\mathbf b}_j),
\end{split}
\end{equation}
where $\phi_{j}:\mathbb{R}^{M_j}\mapsto \mathbb{R}^{N_j}$ is the $j^{th}$ layer of the model. Given ${\textbf{x}}_j \in \mathbb{R}^{M_j}$, the layer applies the affine transform with weights ${\mathbf W}_j \in \mathbb{R}^{N_j \times M_j}$ and biases ${\mathbf b}_j \in \mathbb{R}^{N_j}$ on it, and then pass the resulting vector to the sine nonlinearity.

In our implementation, the concatenated vector $[\textbf{y}_i, \textbf{e}_i]$ is taken as input.
And model weights are shared for all scenes. We use the proposed loss function \eqref{equ:sdfloss} to optimize model weights and shape embeddings.
Note that during training, $\textbf{x}_i$ is sampled from dense ground truth $\Omega_1^\prime$ instead of sparse input $\Omega_1$.
Thus, in a data-driven manner, our generative model can effectively map the shape embedding space to the signed distance space with abundant geometric information.


\subsection{Semantic Extension}
To demonstrate the flexibility of LODE, we extend our method to implicit semantic completion in two ways.

\textbf{Semantic Extension A.} We first semantically segment $V_{\rm{occ}}$ with a sparse CNN. Then a dense CNN is used to predict coarse semantic completion results. Mapping it to our implicit representation using K-Nearest-Neighbor, we get the refined implicit semantic results.

\textbf{Semantic Extension B.} We add a parallel implicit generative head to directly model the implicit semantic label field. Its structure is similar to aforementioned generative model, except that it outputs the probabilities of label classification.

The results are supervised with a cross-entropy loss:
\begin{equation}
\mathcal{L}_{\rm{seg}} = -{\frac 1 {N_{\rm{seg}}}}\sum_{i=1} ^ {N_{\rm{seg}}}\sum_{c=1} ^ {C} y_{i,c} {\rm{log}}(p_{i,c}),
\end{equation}
where $y_{i,c}$ and $p_{i,c}$ are the true and predicted probabilities.
$N_{\rm{seg}}$ points and $C$ categories are considered.

\subsection{Training and Inference}

During training, we randomly sample $N_{\rm{on}}$ on-surface points from $\Omega_1^\prime$ and $N_{\rm{off}}$ off-surface points from $\Omega_3^\prime$, optimizing the whole neural network with loss:
\begin{equation}
\begin{split}
\mathcal{L}_{\rm{total}} = \mathcal{L}_{\rm{LODE}} + \lambda_5\mathcal{L}_{\rm{com}} + \lambda_6\mathcal{L}_{\rm{seg}}.
\end{split}
\end{equation}
Here, $\lambda_5$ and $\lambda_6$ are constant weight parameters. Note that $\lambda_6 = 0$ when the semantic extension is not included.

During inference, we uniformly sample $N_{\rm{inf}}^3$ points from $\Omega_0$ at a specified resolution. And we use a threshold $v_{\rm {th}}$ close to zero to select the points with estimated SDF values smaller than $v_{\rm {th}}$ as  explicit surface points for evaluation.

\begin{figure*}[t]
\centerline{\includegraphics[width=0.8\textwidth]{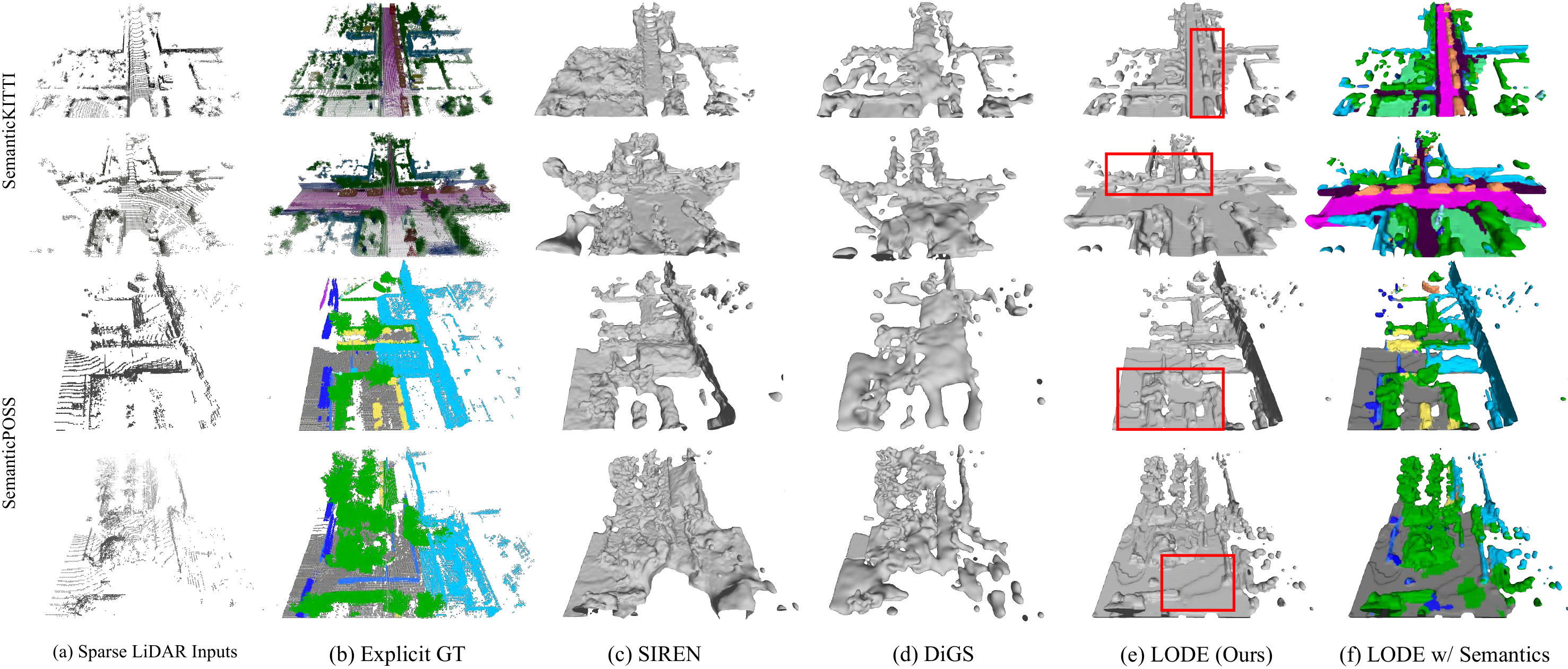}}
\vspace{-0.2cm}
\caption{Qualitative results of Eikonal implicit road scene completion on the SemanticKITTI and SemanticPOSS validation set.}
\label{fig:qualitative}
\vspace{-0.5cm}
\end{figure*}

\section{Experiments}

\textbf{Dataset.} We evaluate the proposed LODE on SemanticKITTI \cite{c2} and SemanticPOSS \cite{pan2020semanticposs}.
There are $22$ sequences (8550 scans) and 6 sequences (2988 scans) of road scene LiDAR data in the two datasets respectively.
Each scan covers a range of 51.2m ahead of the LiDAR, 25.6m to each side, and 6.4m in height.
We follow the official split for training and validation.
\textbf{Metric.} We use the interactions over union (IoU) metric for evaluation.
\textbf{Implementation Details.}
For the discriminative model, we set $D_{\rm{occ}}=256$, $W_{\rm{occ}}=256$, $H_{\rm{occ}}=32$, $m = 5$.
For the generative model, we use $N_{\rm{on}}=N_{\rm{off}}=16000$ and $N_{\rm{inf}}=256$.
For training, we set $\lambda_1=3000$, $\lambda_2=100$, $\lambda_3=100$, $\lambda_4=50$, $\lambda_5=100$ and use the Adam optimizer with an initial learning rate of $10^{-4}$. When the semantic extension is included, $\lambda_6=50$.

\subsection{Scene Completion Effectiveness of LODE}

We compare our LODE with other recent strong Eikonal completion methods on the validation sets of SemanticKITTI and SemanticPOSS. 
The results of these Eikonal methods are obtained by applying them to every LiDAR sweep and taking the average.
In Table.\ref{tab:sc}, the first row shows that directly comparing the input sparse point cloud with completion ground truth yields 10.3\% and 13.0\% IoU.
The existing Eikonal methods improve the IoU to 31.7\% and 40.5\%.
Thanks to the new locally conditioned Eikonal formulation, our approach further improves IoU to 51.2\% and 48.7\%. These results demonstrate the effectiveness of LODE.

\begin{table}[thpb]
\centering
\caption{Scene Completion Results Measured in IoU (\%).}
\vspace{-0.2cm}
\resizebox{0.45\textwidth}{!}{
\begin{tabular}{lccc}
\hline
\multicolumn{1}{c}{Method} & Reference & SemanticKITTI         & SemanticPOSS         \\ \hline
Input                      & - & 10.3                  & 13.0                 \\
SIREN \cite{c10}                     & NeurIPS 2020 & 26.3                  & 36.0                 \\
Fourier Features \cite{tancik2020fourier}          & NeurIPS 2020 & 28.6                  & 30.9                 \\
BACON \cite{lindell2022bacon}                     & CVPR 2022 & 30.5                  & {\ul 40.5}           \\
DiGS \cite{ben2022digs}                      & CVPR 2022 & {\ul 31.7}            & 37.4                 \\
LODE (Ours)                & - & \textbf{51.2 (\textcolor[RGB]{52,168,83}{+19.5})} & \textbf{48.7 (\textcolor[RGB]{52,168,83}{+8.2})} \\ \hline
\end{tabular}}
\label{tab:sc}
\vspace{-0.2cm}
\end{table}

This large improvement is better demonstrated with qualitative results in Fig.~\ref{fig:qualitative}.
Though the existing Eikonal methods are successful for clean synthetic point cloud data uniformly sampled on watertight meshes as shown in \cite{c10}, fitting large-scale outdoor scenes captured by LiDAR (Fig.\ref{fig:qualitative}-a) is much more difficult.
On the one hand, many regions are not sampled thus missing in the point cloud. On the other hand, caused by the mechanism of LiDAR, data sparsity increases with distance and it is extremely sparse at the far end.
These result in the lack of effective boundary values for solving the Eikonal differential equation.
For this reason, as pure generative models, these methods fail to fit road scenes and produce lots of artifacts (Fig.\ref{fig:qualitative}-c,d).
Our LODE, on the contrary, takes data-driven shape priors generated by a strong sparse convolutional network as the dense boundary values and successfully completes the scenes. As shown in Fig.\ref{fig:qualitative}-e and highlighted in red boxes, both occluded and incomplete regions are better reconstructed than other Eikonal methods.

\begin{figure}[thpb]
\centerline{\includegraphics[width=0.4\textwidth]{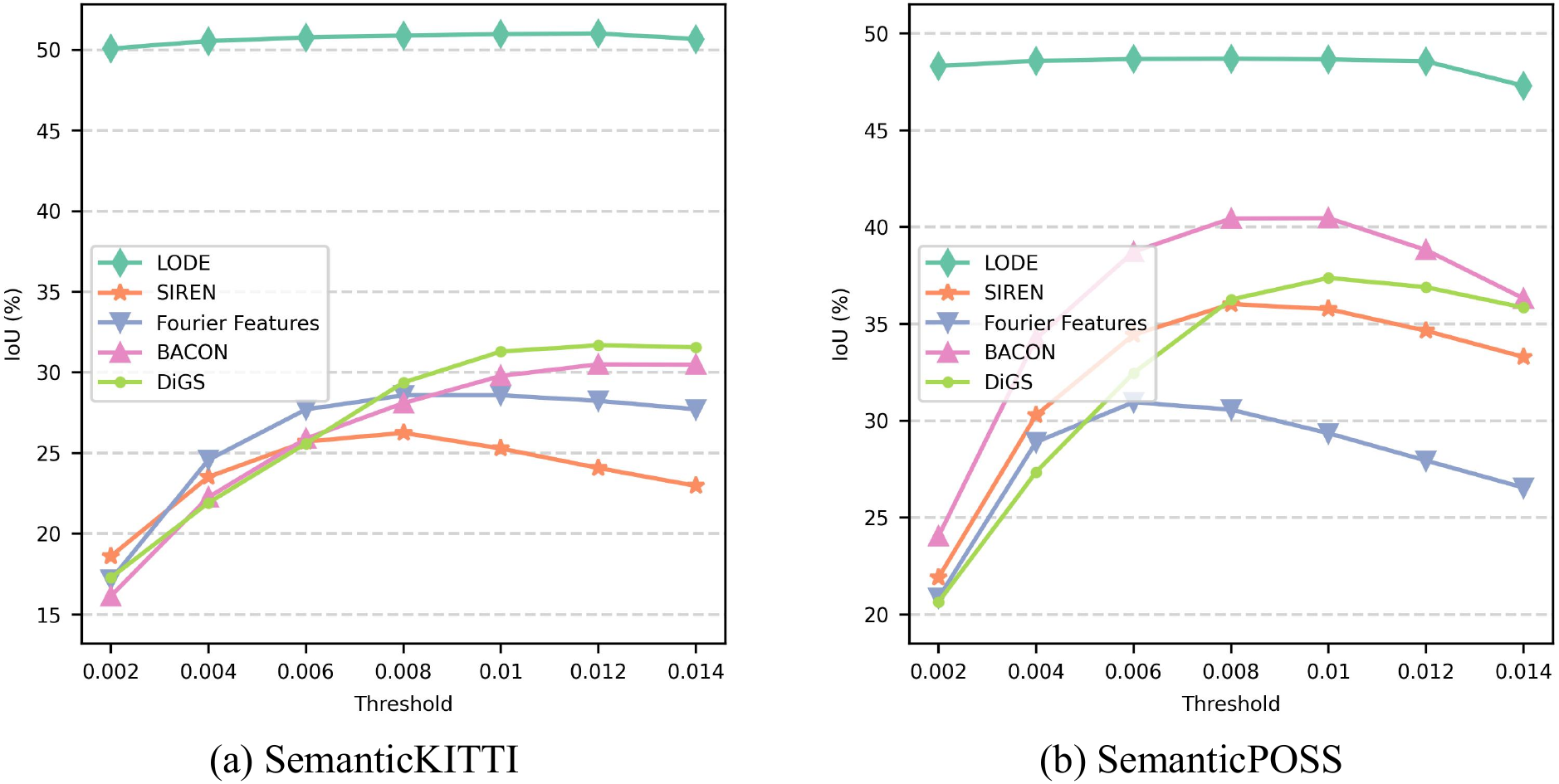}}
\vspace{-0.2cm}
\caption{IoU comparisons under different thresholds.}
\label{fig:IoUcd}
\vspace{-0.2cm}
\end{figure}


In order to show that the significant margins reported in Table.~\ref{tab:sc} are robust to Marching Cubes thresholds, we provide an exhaustive evaluation in Fig.\ref{fig:IoUcd}. It is clear that our method outperforms other methods under all inspected thresholds.


\subsection{Implementation Robustness of LODE}

To better understand LODE and demonstrate its robustness to hyper-parameters in implementation, we provide a series of ablation studies on SemanticKITTI as follows.

\textbf{Discriminative model design.} We investigate two factors: (1) Where to add pruning blocks; (2) Conv layer number in the output block that generates shape embeddings. As shown in Table.\ref{tab:dis_model}, LODE is robust to these design choices.

\textbf{Does generative model capacity matter?} Deeper and wider models usually achieve better results for recognition. To explore whether generative model capacity matters for LODE, we ablate the width, depth, and activations of the MLP. As shown in Table.\ref{tab:gen_model}, different configurations produce similar results. It demonstrates the capacity of generative model is not a performance bottleneck. Interestingly, using ReLU instead of Sine activation only brings a performance drop of 1.75\%. It suggests that in challenging scenarios like ours, using Sine activation is not as critical as in SIREN \cite{c10}.

\textbf{Which dimension of shape volume matters?}
To study which factor is the deciding one for the representation power of the shape volume, we evaluate different shape embedding dimensions and scale sizes. By scale size, we mean the down-sampling ratio $\frac{D_{\rm{occ}}}{D_{\rm{se}}}$. The results are summarized in Table.\ref{tab:DimensionScale}, showing that using shape embeddings of dimension 128 is already capable of representing our scenes well. But increasing scale size leads to a sharp drop of IoU, which reflects the importance of the locality of shape priors.

\begin{table}[htbp]
	\centering
	\begin{minipage}[t]{0.45\linewidth}
		\small\centering
        \caption{Discriminative Model.}
 		\vspace{-0.3cm}
		\resizebox{0.95\textwidth}{!}{
            \begin{tabular}{ccc}
                \hline
                \begin{tabular}[c]{@{}c@{}}Pruning\\ Blocks\end{tabular} & \begin{tabular}[c]{@{}c@{}}Output\\ Block\end{tabular} & IoU (\%)  \\
                \hline
                Last 1  & 2 convs & 49.5          \\
                Last 2  & 2 convs & 49.1          \\
                Last 3  & 2 convs & 50.6          \\
                Last 4  & 2 convs & 51.0          \\
                All & 2 convs & \textbf{51.1} \\
                All & 4 convs & 50.9          \\
                \hline
            \end{tabular}
        }
	    \label{tab:dis_model}
 		\small\centering
        \caption{Generative Model.}
 		\vspace{-0.3cm}
 		\resizebox{0.95\textwidth}{!}{
            \begin{tabular}{cccc}
            \hline
            Width & Depth & Activations & IoU (\%) \\
            \hline
            128   & 4     & Sine        & \textbf{51.0} \\
            256   & 4     & Sine        & 51.0   \\
            512   & 4     & Sine        & 50.9   \\
            256   & 3     & Sine        & 49.6   \\
            256   & 5     & Sine        & 50.9   \\
            256   & 4     & ReLU        & 49.3   \\
            \hline
            \end{tabular}
 	    }
 		\label{tab:gen_model}
	\end{minipage}\hspace{1em}
	\begin{minipage}[t]{0.45\linewidth}
		\small\centering
        \caption{Shape Embedding.}
 		\vspace{-0.3cm}
		\resizebox{0.95\textwidth}{!}{
            \begin{tabular}{ccc}
            \hline
            \begin{tabular}[c]{@{}c@{}}Shape\\ Dimension\end{tabular} & \begin{tabular}[c]{@{}c@{}}Scale\\ Size\end{tabular} & IoU (\%)\\
            \hline
            128&4&50.9\\
            512&4&\textbf{51.2}\\
            256&2&50.3\\
            256&4&51.0\\
            256&8&49.2\\
            256&16&44.8\\
            \hline
            \end{tabular}
        }
		\label{tab:DimensionScale}
 		\small\centering
        \caption{Positional Encoding.}
 		\vspace{-0.3cm}
 		\resizebox{0.95\textwidth}{!}{
            \begin{tabular}{cccc}
            \hline
            \begin{tabular}[c]{@{}c@{}}Positional\\ Encoding\end{tabular} & \begin{tabular}[c]{@{}c@{}}Include\\ xyz\end{tabular} & \begin{tabular}[c]{@{}c@{}}Encoding\\ Level $L$\end{tabular} & IoU (\%) \\
            \hline
            ${\times}$&-&-&40.4\\
            \checkmark&\checkmark&5&40.3\\
            \checkmark&\checkmark&10&51.0\\
            \checkmark&\checkmark&15&50.9\\
            \checkmark&${\times}$&10&\textbf{51.1}\\
            \hline
            \end{tabular}
 	    }
 		\label{tab:AblationPosEncoding}
 	\end{minipage}
\end{table}
\vspace{-0.6cm}
\begin{table}[thpb]
\setlength\tabcolsep{20pt}
\caption{Sampling Strategy.}
\vspace{-0.3cm}
\centering
\begin{tabular}{cc}
\hline
Sample Strategy & IoU (\%) \\
\hline
Trilinear&\textbf{51.0}\\
Nearest&48.1\\
\hline
\end{tabular}
\label{tab:trilinear}
\end{table}

\textbf{Is trilinear sampling necessary?} We justify the necessity of trilinear sampling in our method using Table.~\ref{tab:trilinear}. A trivial nearest neighbor sampling leads to a performance drop of 2.9\%. This is a clear margin that shows the benefit of smoothly interpolating shape embeddings. 

\begin{figure}[thpb]
\centerline{\includegraphics[width=0.4\textwidth]{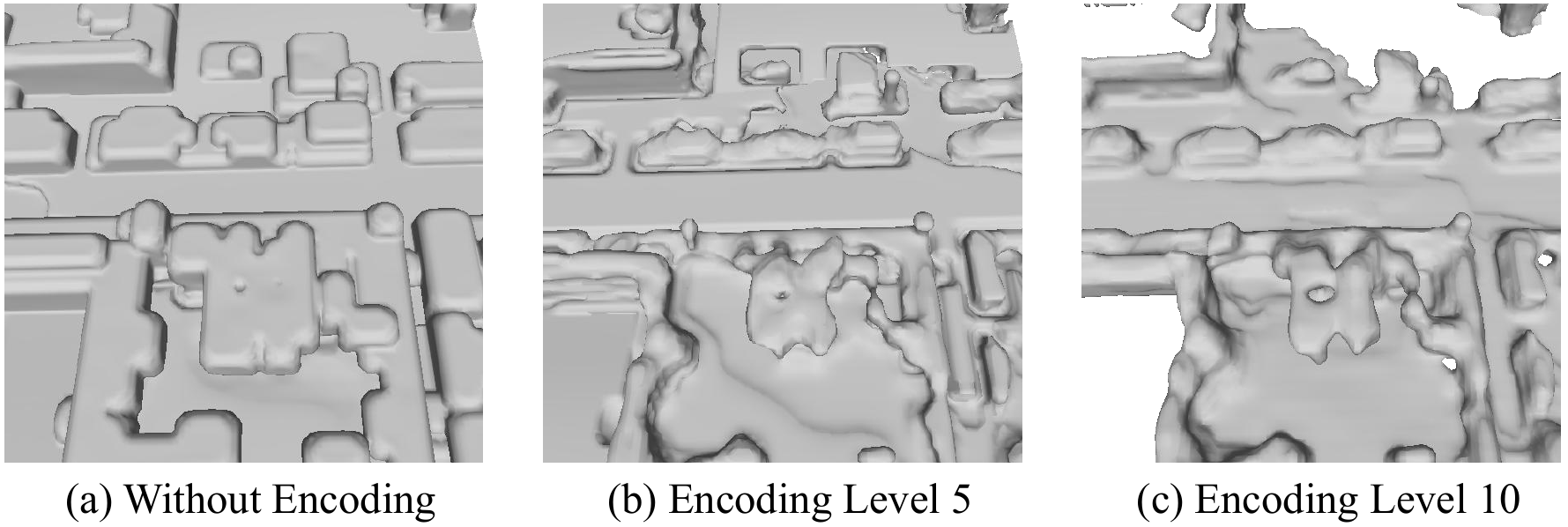}}
\vspace{-0.2cm}
\caption{Qualitative results with different positional encoding strategies.}
\label{fig:encoding_level}
\vspace{-0.3cm}
\end{figure}

\textbf{How to encode positional information?}
The goal of positional encoding is to represent fine geometric details of the scene. We investigate positional encoding levels and whether to concatenate original coordinates.
As shown in Table.\ref{tab:AblationPosEncoding}, when positional encoding is not used or the encoding level is low, the completion IoU decreases dramatically. Through the qualitative results in Fig.~\ref{fig:encoding_level}-a, it is clear that leaving out positional encoding leads to the loss of details.

With these ablations and analyses, we demonstrate the role of each module and the impact of hyper-parameters in detail.
Meanwhile, despite the performance fluctuation under different hyper-parameters, our quantitative results are always better than the existing Eikonal methods shown in the second to the fifth row of Table.I, which demonstrates the effectiveness and robustness of LODE.

\subsection{Flexibility of LODE}
Table.\ref{tab:segmentation} shows semantic scene completion results on the SemanticKITTI validation set, which is evaluated on 19 categories.
With little impact on original completion results, the semantic extension A and B achieve 20.2\% and 18.0\% mIoU, respectively. Although they under-perform the state-of-the-art method JS3C-Net \cite{c8}, our models allow implicit completion and models the signed distance field. Qualitative results shown in Fig.~\ref{fig:qualitative}-f demonstrate faithful semantic implicit completion. Last but not least, we map explicit semantic completion results from JS3C-Net to our implicit completion using K-Nearest-Neighbors, achieving 23.4\% mIoU. These results show the flexibility of LODE as it can be easily extended to provide semantic information.


\begin{table}[thpb]
\centering
\caption{Semantic Scene Completion Results on SemanticKITTI.}
\begin{tabular}{c|cccc}
\hline
Approach  & ext.A & ext.B & JS3C & LODE w/ JS3C \\
\hline
mIoU (\%) & 20.2  & 18.0  & {\ul{22.7}} & \textbf{23.4} \\ 
\hline
\end{tabular}
\label{tab:segmentation}
\vspace{-0.3cm}
\end{table}


\subsection{Other Potential Benefits of LODE}
\begin{figure}[t]
\centerline{\includegraphics[width=0.4\textwidth]{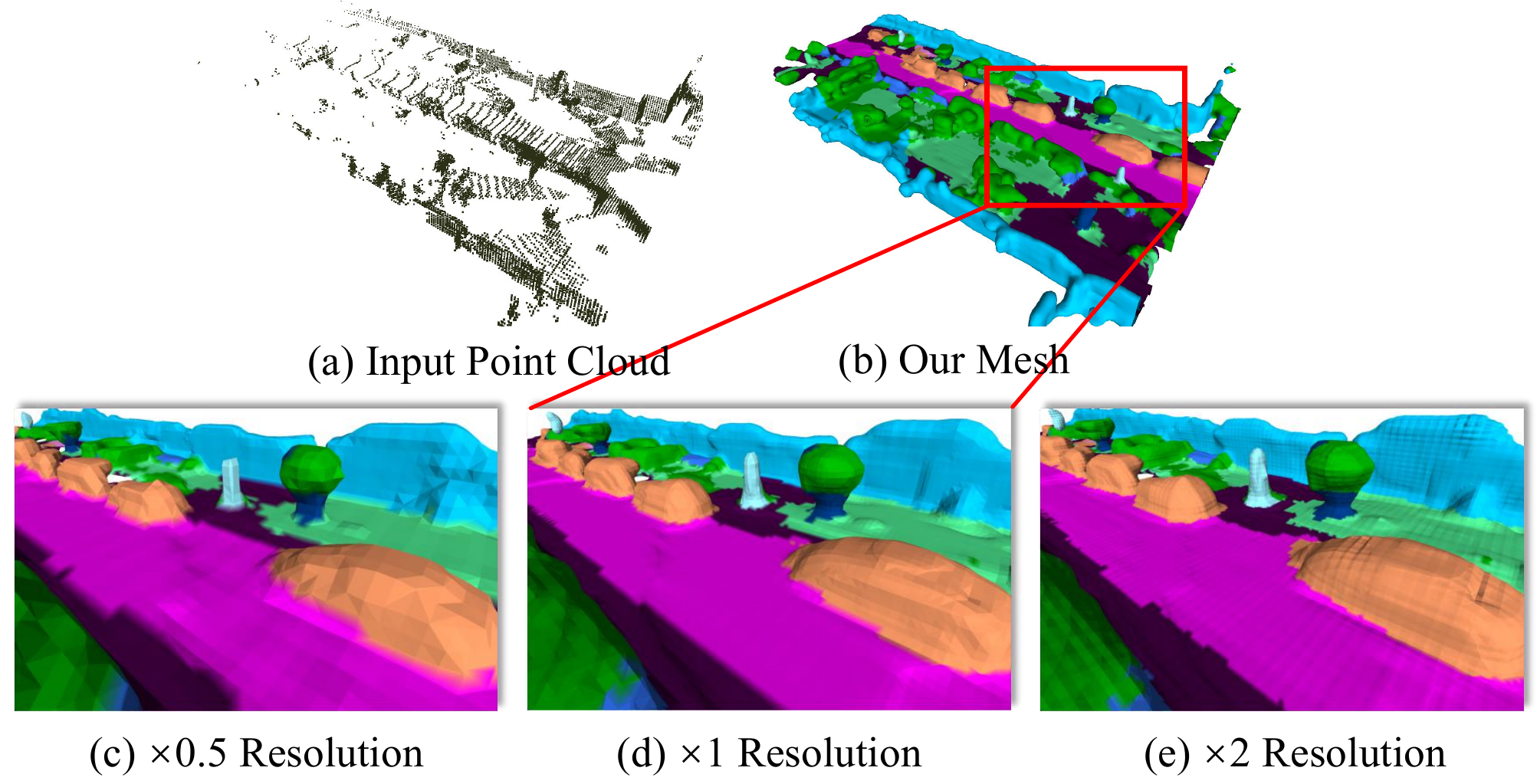}}
\caption{Scene completion results at multiple resolutions.}
\vspace{-0.3cm}
\label{fig:multi_resolution}
\vspace{-0.3cm}
\end{figure}

As illustrated in Fig.\ref{fig:multi_resolution}, with LODE, we can get mesh reconstructions at any resolution, which is due to its \textbf{continuous representation} of the signed distance field.
Meanwhile, as shown in Fig.\ref{fig:teaser} and Fig.\ref{fig:qualitative}, LODE enables the indiscernible LiDAR data to be converted into \textbf{human-friendly visualizations}, which demonstrates its capacity to enhance human understanding of robot-perceived information.
Moreover, the proposed LODE has \textbf{compatibility with implicit planning algorithms} \cite{driess2022learning, li20223d, adamkiewicz2022vision, ha2022deep, pan2022voxfield}, and thus can be leveraged to facilitate downstream robotic manipulation tasks.

\section{Conclusion}
In this study, we propose a novel locally conditioned Eikonal formulation named LODE for implicit scene completion. 
Learned shape embeddings are treated as dense boundary values that constrain signed distance function learning. We implement the formulation as a hybrid neural network combining discriminant and generative models.
The network is trained to implicitly fit road scenes captured by sparse LiDAR point clouds, without accessing exact SDF values in free space.
Large-scale evaluations on SemanticKITTI and SemanticPOSS show that our method outperforms existing Eikonal methods by a large margin.
We also extend the proposed method for semantic implicit completion in two ways, achieving strong qualitative and quantitative results.

\section*{ACKNOWLEDGEMENTS}

This work was sponsored by Tsinghua-Toyota Joint Research Fund (20223930097) and Baidu Inc. through Apollo-AIR Joint Research Center.



\bibliographystyle{IEEEtran}
\bibliography{References}

\end{document}